\documentclass{article}
\usepackage{spconf,amsmath,amssymb,graphicx,hyperref}

\usepackage{booktabs}
\usepackage{multirow}
\usepackage{makecell}
\usepackage[table]{xcolor}
\usepackage{graphicx}
\definecolor{gain}{RGB}{0,125,65}
\definecolor{loss}{RGB}{190,35,35}

\newcommand{\inc}[1]{\textcolor{gain}{\scriptsize$\uparrow$#1}}
\newcommand{\dec}[1]{\textcolor{loss}{\scriptsize$\downarrow$#1}}

\title{INTCORT: Training-Free Spatial Reasoning Enhancement for Vision-Language Models via \underline{In}put \underline{T}ransformations and \underline{Co}nfidence \underline{R}ou\underline{t}ing}
\name{%
Haoran Sun$^{1*}$, Jingqi Xu$^{2*}$, Yanhui Li$^{3}$,
Enci Liu$^{4}$, Kaidi Xu$^{5\dagger}$, Yanwei Liu$^{6\dagger}$%
}

\address{%
$^{1}$The University of Hong Kong,\quad
$^{2}$University of Southern California,\quad
$^{3}$China Telecom\\
$^{4}$Columbia University,\quad
$^{5}$City University of Hong Kong,\quad
$^{6}$Chinese Academy of Sciences\\[2pt]
{\small $^*$Equal contribution.\quad $^\dagger$Corresponding author.}
}

\begin{document}
\fontsize{9.4pt}{10.5pt}\selectfont
\maketitle
\begin{abstract}
Vision-Language Models (VLMs) have demonstrated remarkable capabilities in multimodal tasks, yet they still exhibit poor ability in spatial reasoning. Existing training-dependent and training-free enhancement methods suffer from high computational costs with catastrophic forgetting and internal mechanism interference that compromises general capabilities, respectively. In this work, we first verify two key hypotheses: appropriate geometric image transformation and query-reversal transformation can recover incorrect spatial predictions, and correct predictions exhibit higher relation-token confidence than incorrect ones. Based on these findings, we propose \textbf{INTCORT}, a training-free spatial
reasoning enhancement framework that constructs multiple inference views
through input transformations and aggregates their predictions via
relation-token confidence routing, without modifying the VLM's internal
mechanisms. Experimental results on several commonly-used benchmarks demonstrate that \textbf{INTCORT} substantially improves spatial reasoning accuracy across diverse VLMs, achieving an average improvement of 10.01\% over all models and benchmarks. Compared with prior works, \textbf{INTCORT} achieves superior performance with improvements of up to 25.01\%.
\end{abstract}
\begin{keywords}
Vision-Language Models, Spatial Reasoning, Input Transformations
\end{keywords}
\begin{figure*}[t]
    \centering
    \includegraphics[
        width=0.9\textwidth,
        keepaspectratio
    ]{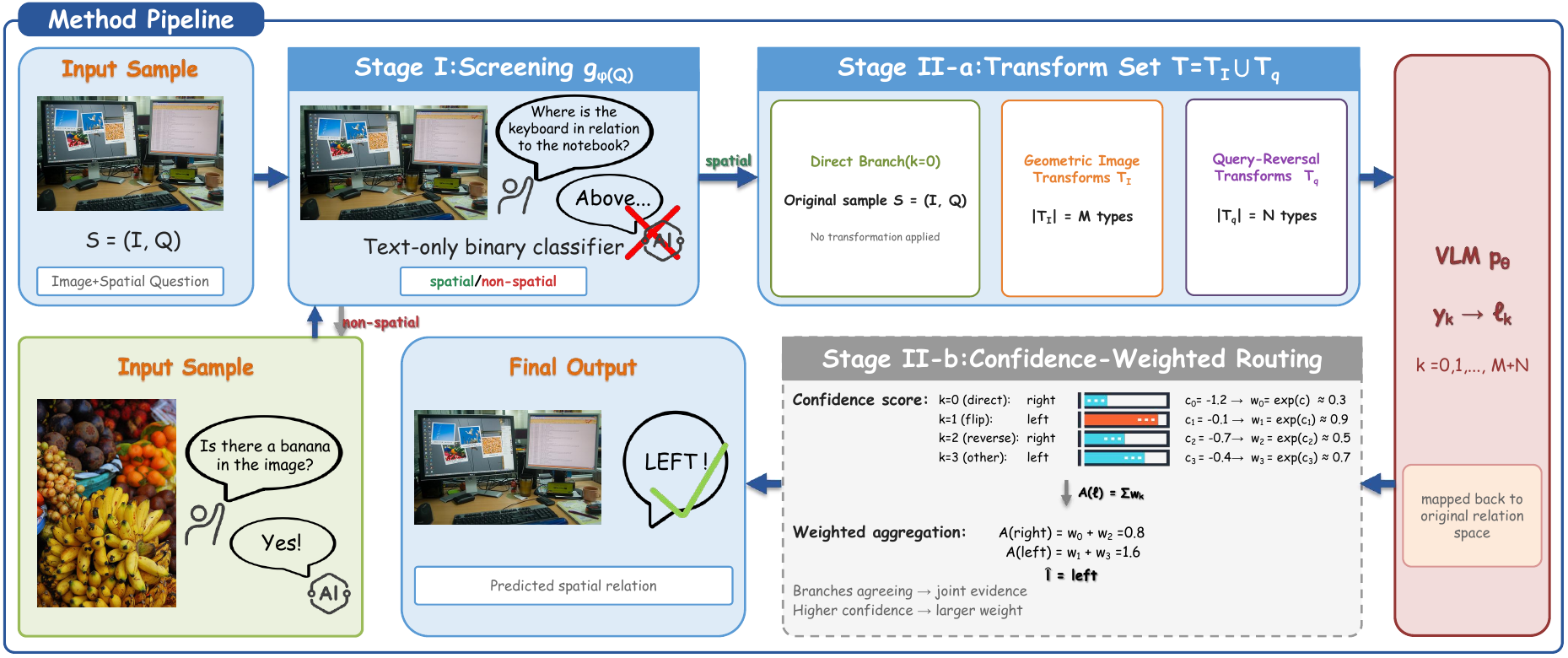}
    \vspace{-1.5mm}
    \caption{Overview of the proposed INTCORT framework.}
    \label{fig:framework}
    \vspace{-2mm}
\end{figure*}

\section{INTRODUCTION}
\label{sec:introduction}

Vision-Language Models (VLMs) have achieved remarkable progress in image captioning, visual question answering, and multimodal dialogue~\cite{alayrac2022flamingo,xu2026redvtp,bai2023qwen}. However, despite their strong general visual understanding capabilities, VLMs still struggle with poor ability in spatial reasoning, such as basic spatial relation queries like ``left/right'' or ``above/below''~\cite{kamath2023s,tong2024cambrian,chen2024spatialvlm}. 
% Accurately comprehending relative positions, orientations, and topological relations among objects is fundamental for embodied intelligence, robot navigation, and complex scene parsing, making this deficiency a critical bottleneck for their deployment in real-world physical interactions~\cite{li2023blip,xu2025hivtp}.
Understanding spatial relations among objects is essential for embodied intelligence, robot navigation, and scene understanding, making this limitation a key obstacle to real-world deployment~\cite{xu2025hivtp}.

To enhance the spatial reasoning capabilities of VLMs, existing methods can be categorized into training-dependent and training-free approaches. For training-dependent methods, they typically use large-scale spatial reasoning data to fine-tune VLMs. For training-free methods, they typically employ internal attention intervention or decoding strategies such as AdaptVis~\cite{chen2025spatial}, DoLa~\cite{chuang2024dola}, and VCD~\cite{leng2024mitigating}. The training-dependent methods incur high computational costs and may cause catastrophic forgetting~\cite{zhai2023investigating}. The training-free methods, as they intervene in the internal mechanisms of VLMs via attention intervention and decoding strategies, may degrade the VLMs' original performance on other reasoning tasks. Therefore, these limitations motivate us to explore a training-free inference strategy
that enhances spatial reasoning without modifying the VLM's internal mechanisms,
thus avoiding compromising its original capabilities.

In this work, our key intuition is that an incorrect prediction from the
original input does not necessarily indicate a lack of spatial reasoning
ability, as VLM predictions may depend on the input view.
Geometric image transformations and query reversal provide alternative
inference views from which incorrect direct predictions may be recovered.
Based on this intuition, we formulate and validate two hypotheses:
appropriate transformations can recover incorrect spatial relation predictions,
and correct predictions tend to exhibit higher relation-token confidence than
incorrect ones. Motivated by these findings, we propose \textbf{INTCORT}, a two-stage training-free framework based on input
transformations and confidence routing. Specifically,
\textbf{INTCORT} first uses a lightweight language model to identify spatial
queries and avoid unnecessary transformations for non-spatial inputs.
For spatial queries, it then constructs multiple transformed inference branches,
maps the predictions produced by the VLMs back to the original answer space,
and aggregates them using relation-token confidence.
This multi-view evidence aggregation improves spatial reasoning reliability
without modifying the VLM's internal mechanisms.

To evaluate the effectiveness of \textbf{INTCORT}, we conduct extensive
experiments on seven spatial reasoning benchmarks\cite{tong2024cambrian,chen2025spatial,kamath2023s,krishna2017visual,lin2014microsoft} across six mainstream VLMs,
demonstrating consistent improvements over corresponding base models and existing
training-free methods. Our main contributions are summarized as follows: 1) We reveal that spatial reasoning failures in VLMs can often be recovered through alternative inference views and propose \textbf{INTCORT}, a training-free spatial reasoning enhancement framework that operates exclusively at the input side via multi-view transformations and confidence routing without modifying the VLM's internal mechanisms. 2) We verify that geometric transformations can recover incorrect spatial predictions and that correct transformed predictions tend to have higher relation-token confidence scores than incorrect direct predictions. 3) We validate the superiority and complementarity of \textbf{INTCORT} across seven commonly-used spatial reasoning benchmarks and six mainstream VLMs, without degrading performance on general benchmarks.

\section{PROPOSED METHOD}
\label{sec:method}

Spatial reasoning requires VLMs to understand the
spatial arrangement of entities in a visual scene and express the corresponding
relations in language. 
Formally, given an image \(I\) and a spatial reasoning
question \(q\), a VLM parameterized by \(\theta\) generates a response \(y\)
according to \(y \sim p_{\theta}(\cdot \mid I,q)\).
% Formally, given an image \(I\) and a spatial reasoning
% question \(q\), a VLM parameterized by \(\theta\) generates a response \(y\)
% according to
% \begin{equation}
%     y \sim p_{\theta}(\cdot \mid I,q).
%     \label{eq:spatial_reasoning_task}
% \end{equation}
Despite their strong visual understanding capabilities, current VLMs remain
unreliable in spatial reasoning.
In this work, we focus on two representative types of spatial reasoning,
namely the position of an entity with respect to the image and the spatial
relation between two entities in the image.

\subsection{Recovering Spatial Predictions via Transformations}
\label{sec:evidence}

\textbf{Motivation.}
Intuitively, directly answering a spatial reasoning question from the original input may lead to an incorrect relation prediction.
Geometric image transformations and query reversal provide complementary
inference views of the same underlying spatial relation. These views may help
the VLM better perceive and compare spatial cues, allowing it to produce the
correct relation when direct inference fails.
Based on this intuition, we formulate two hypotheses.
First, appropriate transformations can recover incorrect spatial predictions.
Second, correct predictions are expected to exhibit higher confidence on the generated relation token than incorrect ones.
To validate these hypotheses, we conduct experiments with Qwen2-VL-7B on the
Controlled-B and CV-Bench-R benchmarks.
The detailed experimental setup is provided in Section~\ref{sec:experiments}.

Specifically, we define a geometric image transformation set
\(\mathcal{T}_{I}\) with \(|\mathcal{T}_{I}|=M\), including transformations
such as horizontal and vertical flips.
We further define a query-reversal transformation set
\(\mathcal{T}_{q}\) with \(|\mathcal{T}_{q}|=N\), where each transformation
reverses the query direction by swapping the target and reference entities.
For example, ``Where is the mug in relation to the knife?'' is transformed
into ``Where is the knife in relation to the mug?'', with \(N=1\) for this example.
Combining the two sets, we define the complete transformation set as
\(\mathcal{T}=\mathcal{T}_{I}\cup\mathcal{T}_{q}\), with
\(|\mathcal{T}|=M+N\).
% Combining the two sets, we define the complete transformation set as
% \begin{equation}
%     \mathcal{T}
%     =
%     \mathcal{T}_{I}\cup\mathcal{T}_{q},
%     \qquad
%     |\mathcal{T}|=M+N.
%     \label{eq:complete_transform}
% \end{equation}

% For each sample \(S=(I,Q)\) in the Controlled-B and CV-Bench Relation
% benchmarks, we first perform direct inference using the VLM and collect the
% incorrectly predicted samples into
% \begin{equation}
%     \mathcal{S}_{\mathrm{err}}
%     =
%     \left\{S^{(i)}\right\}_{i=1}^{n_{\mathrm{err}}},
%     \qquad
%     |\mathcal{S}_{\mathrm{err}}|=n_{\mathrm{err}},
%     \label{eq:error_set}
% \end{equation}
% where \(n_{\mathrm{err}}\) is the number of samples incorrectly predicted by
% direct inference.
For each sample \(S=(I,Q)\) in the Controlled-B and CV-Bench-R
benchmarks, we first perform direct inference using the VLM and collect the
incorrectly predicted samples into
\(\mathcal{S}_{\mathrm{err}}=\{S^{(i)}\}_{i=1}^{n_{\mathrm{err}}}\), where
\(|\mathcal{S}_{\mathrm{err}}|=n_{\mathrm{err}}\) and \(n_{\mathrm{err}}\)
denotes the number of samples incorrectly predicted by direct inference.
% For each erroneous sample \(S^{(i)} \in \mathcal{S}_{\mathrm{err}}\), we apply
% every transformation \(t_k \in \mathcal{T}\) to obtain
% \begin{equation}
%     S_k^{(i)}
%     =
%     t_k\left(S^{(i)}\right),
%     \qquad
%     k=1,\ldots,M+N,
%     \label{eq:transformed_sample}
% \end{equation}
% where \(S_k^{(i)}\) denotes the transformed sample obtained by applying the
% \(k\)-th transformation to \(S^{(i)}\).
For each erroneous sample \(S^{(i)} \in \mathcal{S}_{\mathrm{err}}\), we apply
every transformation \(t_k \in \mathcal{T}\) to obtain
\(S_k^{(i)} = t_k\!\left(S^{(i)}\right)\), where
\(k=1,\ldots,M+N\) and \(S_k^{(i)}\) denotes the sample generated by applying
the \(k\)-th transformation to \(S^{(i)}\).
% We then define the set of transformed samples corresponding to
% \(S^{(i)}\) as
% \begin{equation}
%     \mathcal{S}_{\mathrm{trans}}^{(i)}
%     =
%     \left\{
%     S_k^{(i)}
%     \right\}_{k=1}^{M+N},
%     \qquad
%     \left|
%     \mathcal{S}_{\mathrm{trans}}^{(i)}
%     \right|
%     =
%     M+N.
%     \label{eq:transformed_sample_set}
% \end{equation}
We then define the transformed sample set corresponding to \(S^{(i)}\) as
\(\mathcal{S}_{\mathrm{trans}}^{(i)}
=\{S_k^{(i)}\}_{k=1}^{M+N}\), with
\(|\mathcal{S}_{\mathrm{trans}}^{(i)}|=M+N\).

Next, we perform VLM inference on every transformed sample in
\(\mathcal{S}_{\mathrm{trans}}^{(i)}\).
An erroneous sample \(S^{(i)}\) is considered recovered if at least
one of its \(M+N\) transformed samples produces the correct relation
prediction after relation alignment.
We denote the total number of recovered samples by \(n_{\mathrm{rec}}\) and define the recovery ratio as \(\rho_{\mathrm{rec}} = n_{\mathrm{rec}} / n_{\mathrm{err}}\).
For each recovered sample, we further record the confidence score
\(c_{\mathrm{rec}}^{(i)}\) of the correct transformed prediction and compare it with the confidence score
\(c_{\mathrm{dir}}^{(i)}\) of the incorrect direct prediction.
We denote by \(n_{\mathrm{conf}}\) the number of recovered samples satisfying
\(c_{\mathrm{rec}}^{(i)} > c_{\mathrm{dir}}^{(i)}\), and define the confidence superiority ratio as
\(\rho_{\mathrm{conf}} = n_{\mathrm{conf}} / n_{\mathrm{rec}}\).

% \begin{table}[t]
% \centering
% \scriptsize
% \renewcommand{\arraystretch}{0.65}
% \setlength{\aboverulesep}{0pt}
% \setlength{\belowrulesep}{0pt}

% \resizebox{0.95\columnwidth}{!}{%
% \begin{tabular}{llccccc}
% \toprule
% Model
% & Benchmark
% & $n_{\mathrm{err}}$
% & $n_{\mathrm{rec}}$
% & $n_{\mathrm{conf}}$
% & $\rho_{\mathrm{rec}}$
% & $\rho_{\mathrm{conf}}$ \\
% \midrule

% Qwen2-VL-7B
% & \makecell[l]{Controlled-B\\+ CV-Bench-R}
% & 141
% & 130
% & 107
% & 92.20\%
% & 82.31\% \\

% \bottomrule
% \end{tabular}%
% }

% \caption{Analysis of recoverable spatial predictions and confidence.}
% \label{tab:motivation}
% \end{table}

\begin{table}[t]
\centering

{\footnotesize % 约 8pt
\renewcommand{\arraystretch}{1}
\setlength{\tabcolsep}{1.5pt}
\setlength{\aboverulesep}{0.5pt}
\setlength{\belowrulesep}{0.5pt}

\begin{tabular}{@{}llccccc@{}}
\toprule
Model
& Benchmark
& $n_{\mathrm{err}}$
& $n_{\mathrm{rec}}$
& $n_{\mathrm{conf}}$
& $\rho_{\mathrm{rec}}$
& $\rho_{\mathrm{conf}}$ \\
\midrule

Qwen2-VL-7B
& \makecell[l]{Controlled-B\\+ CV-Bench-R}
& 141
& 130
& 107
& 92.20\%
& 82.31\% \\

\bottomrule
\end{tabular}
}

\caption{Analysis of recoverable spatial predictions and confidence.}
\label{tab:motivation}
\end{table}

As shown in Table~\ref{tab:motivation}, the recovery ratio
\(\rho_{\mathrm{rec}}\) reaches \(92.20\%\), indicating that, for most samples
incorrectly answered by direct inference, at least one transformed view can
recover the correct spatial relation.
Moreover, the confidence superiority ratio \(\rho_{\mathrm{conf}}\) reaches
\(82.31\%\), showing that the correct transformed prediction has a higher
relation-token confidence than the incorrect direct prediction in the
majority of recovered samples.
These results provide support for our two hypotheses and motivate
the use of transformed inference views together with relation-token confidence
for spatial reasoning.

\begin{table*}[t]
\centering
\scriptsize

\renewcommand{\arraystretch}{1}
\setlength{\aboverulesep}{0.5pt}
\setlength{\belowrulesep}{0.5pt}

\resizebox{0.97\textwidth}{!}{%
\begin{tabular}{cc*{7}{c}}
\toprule
\multicolumn{1}{c}{\textbf{Model}} &
\multicolumn{1}{c}{\textbf{Method}} &
VG-one &
VG-two &
COCO-one &
COCO-two &
CV-Bench-R &
Controlled-A &
Controlled-B \\
\midrule

\multirow{5}{*}{\makecell[c]{Qwen-VL\\Chat}}
& Original model
& 41.81 & 36.05 & 44.72 & 57.10 & 54.42 & 64.16 & 63.11 \\
& +AdaptVis
& 51.40\inc{9.59} & 46.35\inc{10.30}
& 52.78\inc{8.06} & 57.39\inc{0.29}
& 55.01\inc{0.59} & 64.97\inc{0.81}
& 63.41\inc{0.30} \\
& +DoLa
& 41.49\dec{0.32} & 37.34\inc{1.29}
& 45.78\inc{1.06} & 57.36\inc{0.26}
& 54.61\inc{0.19} & 64.37\inc{0.21}
& 62.80\dec{0.31} \\
& +VCD
& 39.96\dec{1.85} & 38.20\inc{2.15}
& 47.06\inc{2.34} & 56.42\dec{0.68}
& 55.19\inc{0.77} & 63.55\dec{0.61}
& 62.28\dec{0.83} \\
\rowcolor{gray!15}
& \multicolumn{1}{c}{\textbf{+INTCORT}}
& \textbf{56.68}\inc{14.87} & \textbf{57.94}\inc{21.89}
& \textbf{53.00}\inc{8.28} & \textbf{64.49}\inc{7.39}
& \textbf{57.69}\inc{3.27} & \textbf{65.12}\inc{0.96}
& \textbf{64.16}\inc{1.05} \\

\midrule

\multirow{5}{*}{\makecell[c]{Qwen2-VL\\7B}}
& Original model
& 73.92 & 57.51 & 69.35 & 75.57 & 78.27 & 97.89 & 91.46 \\
& +AdaptVis
& 74.14\inc{0.22} & 67.09\inc{9.58}
& 70.44\inc{1.09} & 77.55\inc{1.98}
& 78.85\inc{0.58} & 97.59\dec{0.30}
& 91.81\inc{0.35} \\
& +DoLa
& 79.09\inc{5.17} & 66.18\inc{8.67}
& 69.46\inc{0.11} & 76.42\inc{0.85}
& 78.46\inc{0.19} & 97.48\dec{0.41}
& 92.22\inc{0.76} \\
& +VCD
& 76.51\inc{2.59} & 58.00\inc{0.49}
& 66.24\dec{3.11} & 74.43\dec{1.14}
& 79.42\inc{1.15} & 96.69\dec{1.20}
& 91.16\dec{0.30} \\
\rowcolor{gray!15}
& \multicolumn{1}{c}{\textbf{+INTCORT}}
& \textbf{79.20}\inc{5.28} & \textbf{68.67}\inc{11.16}
& \textbf{70.91}\inc{1.56} & \textbf{79.26}\inc{3.69}
& \textbf{82.89}\inc{4.62} & \textbf{99.10}\inc{1.21}
& \textbf{99.09}\inc{7.63} \\

\midrule

\multirow{5}{*}{\makecell[c]{Qwen2.5-VL\\7B}}
& Original model
& 61.85 & 58.37 & 68.74 & 77.84 & 88.46 & 95.48 & 96.04 \\
& +AdaptVis
& 62.61\inc{0.76} & 59.66\inc{1.29}
& 69.69\inc{0.95} & 78.69\inc{0.85}
& 89.23\inc{0.77} & 95.48
& 96.95\inc{0.91} \\
& +DoLa
& 62.18\inc{0.33} & 59.66\inc{1.29}
& 67.90\dec{0.84} & 76.45\dec{1.39}
& 89.38\inc{0.92} & 86.75\dec{8.73}
& 95.20\dec{0.84} \\
& +VCD
& 62.59\inc{0.74} & 59.02\inc{0.65}
& 67.41\dec{1.33} & 78.69\inc{0.85}
& 89.04\inc{0.58} & 91.57\dec{3.91}
& 96.70\inc{0.66} \\
\rowcolor{gray!15}
& \multicolumn{1}{c}{\textbf{+INTCORT}}
& \textbf{73.81}\inc{11.96} & \textbf{72.96}\inc{14.59}
& \textbf{70.80}\inc{2.06} & \textbf{81.82}\inc{3.98}
& \textbf{90.77}\inc{2.31} & \textbf{96.08}\inc{0.60}
& \textbf{99.39}\inc{3.35} \\

\midrule

\multirow{5}{*}{\makecell[c]{Qwen3-VL\\8B}}
& Original model
& 69.94 & 76.39 & 72.36 & 81.25 & 91.15 & \textbf{99.40} & 97.26 \\
& +AdaptVis
& 72.21\inc{2.27} & 77.67\inc{1.28}
& 71.69\dec{0.67} & 81.53\inc{0.28}
& 91.15 & \textbf{99.40}
& 97.48\inc{0.22} \\
& +DoLa
& 70.61\inc{0.67} & 77.54\inc{1.15}
& \textbf{72.97}\inc{0.61} & 80.11\dec{1.14}
& 91.15 & 98.49\dec{0.91}
& 97.56\inc{0.30} \\
& +VCD
& 71.18\inc{1.24} & 77.39\inc{1.00}
& 70.52\dec{1.84} & 81.25
& 91.35\inc{0.20} & 97.59\dec{1.81}
& 97.56\inc{0.30} \\
\rowcolor{gray!15}
& \multicolumn{1}{c}{\textbf{+INTCORT}}
& \textbf{79.31}\inc{9.37} & \textbf{83.69}\inc{7.30}
& \textbf{72.97}\inc{0.61} & \textbf{82.95}\inc{1.70}
& \textbf{94.42}\inc{3.27} & \textbf{99.40}
& \textbf{100.00}\inc{2.74} \\

\midrule

\multirow{5}{*}{\makecell[c]{Molmo\\7B}}
& Original model
& 28.77 & 61.37 & 61.18 & 63.92 & 50.77 & 69.28 & 73.78 \\
& +AdaptVis
& 28.99\inc{0.22} & 62.23\inc{0.86}
& 61.29\inc{0.11} & 64.49\inc{0.57}
& 50.96\inc{0.19} & \textbf{69.88}\inc{0.60}
& 73.78 \\
& +DoLa
& 28.99\inc{0.22} & 63.09\inc{1.72}
& 60.85\dec{0.33} & 64.77\inc{0.85}
& 50.77 & 69.28
& 74.39\inc{0.61} \\
& +VCD
& 30.71\inc{1.94} & 64.81\inc{3.44}
& 58.06\dec{3.12} & 64.49\inc{0.57}
& 51.15\inc{0.38} & 65.66\dec{3.62}
& 80.79\inc{7.01} \\
\rowcolor{gray!15}
& \multicolumn{1}{c}{\textbf{+INTCORT}}
& \textbf{33.30}\inc{4.53} & \textbf{76.39}\inc{15.02}
& \textbf{63.29}\inc{2.11} & \textbf{69.03}\inc{5.11}
& \textbf{52.88}\inc{2.11} & \textbf{71.69}\inc{2.41}
& \textbf{83.54}\inc{9.76} \\

\midrule

\multirow{5}{*}{\makecell[c]{LLaVA-NeXT\\7B}}
& Original model
& 35.34 & 10.30 & 63.96 & 48.01 & 60.96 & 51.51 & 64.02 \\
& +AdaptVis
& 48.06\inc{12.72} & 13.30\inc{3.00}
& 64.63\inc{0.67} & \textbf{50.57}\inc{2.56}
& 61.92\inc{0.96} & \textbf{97.29}\inc{45.78}
& \textbf{77.74}\inc{13.72} \\
& +DoLa
& 35.67\inc{0.33} & 8.15\dec{2.15}
& 64.07\inc{0.11} & 47.73\dec{0.28}
& 60.96 & 51.51
& 64.02 \\
& +VCD
& 37.47\inc{2.13} & 14.74\inc{4.44}
& 64.07\inc{0.11} & 48.58\inc{0.57}
& 61.34\inc{0.38} & 63.13\inc{11.62}
& 66.03\inc{2.01} \\
\rowcolor{gray!15}
& \multicolumn{1}{c}{\textbf{+INTCORT}}
& \textbf{48.92}\inc{13.58} & \textbf{17.60}\inc{7.30}
& \textbf{65.24}\inc{1.28} & 49.43\inc{1.42}
& \textbf{70.00}\inc{9.04} & 53.11\inc{1.60}
& 64.94\inc{0.92} \\

\bottomrule
\end{tabular}%
}

\caption{Comparison with existing training-free inference methods.
All values are accuracies (\%). }
\label{tab:main_results}
\end{table*}

% \begin{table}[t]
% \centering
% \tiny
% \renewcommand{\arraystretch}{0.62}
% \setlength{\aboverulesep}{0pt}
% \setlength{\belowrulesep}{0pt}

% \resizebox{0.88\columnwidth}{!}{%
% \begin{tabular}{lcccc}
% \toprule
% Benchmark
% & LLaVA-NeXT 7B
% & +AdaptVis
% & +INTCORT
% & \makecell{+AdaptVis\\+INTCORT} \\
% \midrule

% VG-one
% & 35.34
% & 48.06
% & 48.92
% & \textbf{62.02} \\

% VG-two
% & 10.30
% & 13.30
% & 17.60
% & \textbf{22.60} \\

% COCO-one
% & 63.96
% & 64.63
% & 65.24
% & \textbf{67.68} \\

% COCO-two
% & 48.01
% & 50.57
% & 49.43
% & \textbf{52.85} \\

% CV-Bench-R
% & 60.96
% & 61.92
% & 70.00
% & \textbf{72.77} \\

% Controlled-A
% & 51.51
% & 97.29
% & 53.11
% & \textbf{98.99} \\

% Controlled-B
% & 64.02
% & 77.74
% & 64.94
% & \textbf{84.01} \\

% \bottomrule
% \end{tabular}%
% }

% \caption{Complementarity analysis.All are accuracies (\%).}
% \label{tab:complementarity}
% \end{table}

\begin{table}[t]
\centering

{\fontsize{8}{9.2}\selectfont
\renewcommand{\arraystretch}{1}
\setlength{\tabcolsep}{2pt}
\setlength{\aboverulesep}{0.5pt}
\setlength{\belowrulesep}{0.5pt}

\begin{tabular}{lcccc}
\toprule
Benchmark
& LLaVA-NeXT 7B
& +AdaptVis
& +INTCORT
& \makecell{+AdaptVis\\+INTCORT} \\
\midrule

VG-one
& 35.34
& 48.06
& 48.92
& \textbf{62.02} \\

VG-two
& 10.30
& 13.30
& 17.60
& \textbf{22.60} \\

COCO-one
& 63.96
& 64.63
& 65.24
& \textbf{67.68} \\

COCO-two
& 48.01
& 50.57
& 49.43
& \textbf{52.85} \\

CV-Bench-R
& 60.96
& 61.92
& 70.00
& \textbf{72.77} \\

Controlled-A
& 51.51
& 97.29
& 53.11
& \textbf{98.99} \\

Controlled-B
& 64.02
& 77.74
& 64.94
& \textbf{84.01} \\

\bottomrule
\end{tabular}%
}

\caption{Complementarity analysis. All are accuracies (\%).}
\label{tab:complementarity}
\end{table}

\subsection{INTCORT Framework}
\label{sec:INTCORT_framework}

Based on the two verified hypotheses, we propose \textsc{INTCORT}, whose overall
framework is illustrated in Fig.~\ref{fig:framework}.
Given an input sample \(S=(I,Q)\), \textsc{INTCORT} performs inference in two
stages. The first stage determines whether the input question requires spatial
reasoning, while the second stage constructs transformed inference views and
aggregates their relation predictions based on relation-token confidence.

\textbf{Stage I: Spatial Query Screening.}
This stage determines whether the input question requires spatial reasoning, as applying spatial transformations to non-spatial questions may introduce irrelevant variations.
Specifically, we feed the input question \(Q\) into a lightweight language
model \(g_{\phi}\), which performs binary classification such that
\(g_{\phi}(Q)\in\{\text{spatial},\text{non-spatial}\}\).
If \(g_{\phi}(Q)=\text{non-spatial}\), we directly query the VLM using the
original input; otherwise, the sample is passed to the second stage of
\textsc{INTCORT}. This screening step prevents unnecessary transformations on non-spatial tasks and preserves the original capability of VLMs.
% Specifically, we feed the input question \(Q\) into a lightweight language
% model \(g_{\phi}\), which performs binary classification to determine whether
% \(Q\) requires spatial reasoning:
% \begin{equation}
%     g_{\phi}(Q)
%     \in
%     \{\text{spatial},\text{non-spatial}\}.
%     \label{eq:spatial_screening}
% \end{equation}
% If \(g_{\phi}(Q)=\text{non-spatial}\), we directly query the VLM using the
% original input.
% If \(g_{\phi}(Q)=\text{spatial}\), the sample is passed to the second stage
% of \textsc{INTCORT}.

\textbf{Stage II: Transformation and Confidence Routing.}
For a spatial reasoning question, we first perform direct inference on the
original input and construct a set of transformed inputs using the
transformation set \(\mathcal{T}\) defined in Section~\ref{sec:evidence}.
We denote the direct inference branch by \(k=0\) and the \(M+N\) transformed branches by
\(k=1,\ldots,M+N\).
Each branch is independently processed by the VLM to generate a response
\(y_k\), from which we extract the predicted spatial relation label
\(\ell_k\). We then map \(\ell_k\) back to the spatial relation space of the original
input and denote the mapped relation label by \(\ell_k^{\star}\), with
\(\ell_0^{\star}=\ell_0\) for the direct branch.
We use the confidence of the generated relation tokens to measure the
reliability of each branch prediction.
Let \(\mathcal{P}_k\) denote the set of token positions in \(y_k\) corresponding
to the tokens that constitute \(\ell_k\).
We define the relation-token confidence score as
\begin{equation}
    c_k
    =
    \min_{t\in\mathcal{P}_k}
    \log p_{\theta}
    \left(
        y_{k,t}\mid y_{k,<t}, I_k, Q_k
    \right),
    \label{eq:confidence_score}
\end{equation}
where \(y_{k,t}\) denotes the \(t\)-th generated token.
For a relation label \(\ell_k\) consisting of multiple tokens, the minimum
token log-probability is used as its confidence score.
We further convert \(c_k\) into a positive aggregation weight as \(w_k=\exp(c_k)\). For each candidate relation label \(\ell\), we aggregate the weights of all
branches whose mapped predictions correspond to \(\ell\) as
\begin{equation}
    A(\ell)
    =
    \sum_{k=0}^{M+N}
    \mathbb{I}\left[\ell_k^{\star}=\ell\right] w_k,
    \label{eq:relation_aggregation}
\end{equation}
where \(\mathbb{I}[\cdot]\) denotes the indicator function.
The final relation prediction is selected as
\(\widehat{\ell}=\arg\max_{\ell} A(\ell)\).
This confidence-weighted aggregation favors relation predictions with broader
branch support and higher relation-token confidence.
% The final relation prediction is then selected as
% \begin{equation}
%     \widehat{\ell}
%     =
%     \arg\max_{\ell} A(\ell).
%     \label{eq:final_relation}
% \end{equation}
% This confidence-weighted aggregation favors relation predictions with broader branch support and higher relation-token confidence.

\begin{table}[t]
\centering

{\fontsize{8}{9.2}\selectfont
\renewcommand{\arraystretch}{1}
\setlength{\tabcolsep}{1.5pt}
\setlength{\aboverulesep}{0.5pt}
\setlength{\belowrulesep}{0.5pt}

\begin{tabular}{@{}lcccc@{}}
\toprule
Benchmark
& \makecell[c]{INTCORT\\(Original)}
& \makecell[c]{w/ Qwen2-0.5B-\\Instruct Router}
& \makecell[c]{w/ Mean\\Confidence}
& \makecell[c]{w/ Max\\Confidence} \\
\midrule

Controlled-A
& \textbf{99.10}
& 98.09
& 97.29
& 93.37 \\

VG-one
& \textbf{79.20}
& 73.93
& 78.97
& 79.09 \\

\bottomrule
\end{tabular}%
}

\caption{Ablation results on Qwen2-VL-7B. All values are accuracies (\%).}
\label{tab:ablation}
\end{table}

\section{EXPERIMENTS}
\label{sec:experiments}
We evaluate \textsc{INTCORT} on seven widely used spatial reasoning benchmarks
using Qwen-VL-Chat~\cite{bai2023qwen}, Qwen2-VL-7B~\cite{wang2024qwen2},
Qwen2.5-VL-7B~\cite{bai2025qwen25vltechnicalreport}, Qwen3-VL-8B~\cite{bai2025qwen3},
Molmo-7B~\cite{deitke2025molmo}, and LLaVA-NeXT-7B~\cite{liu2024llavanext}, and compare it
with state-of-the-art training-free methods, including
AdaptVis~\cite{chen2025spatial}, DoLa~\cite{chuang2024dola}, and VCD~\cite{leng2024mitigating}.
We use accuracy as the evaluation metric.
For the spatial query classifier \(g_{\phi}\), we use Qwen2.5-1.5B-Instruct.
The geometric image transformation set is fixed as
\(\mathcal{T}_{I}=\{H,V\}\), where \(H\) and \(V\) denote horizontal and
vertical flips, respectively.
All experiments are conducted using NVIDIA A800 GPUs.

\subsection{Benchmarks}
\label{sec:benchmarks}
The benchmark suite covers both controlled and real-world scenes.
\textbf{Controlled-A} and \textbf{Controlled-B} are the two controlled
benchmarks introduced in AdaptVis~\cite{chen2025spatial}.
They contain clean-background images with two objects.
Controlled-A consists of one large object and one small object, whereas
Controlled-B contains two small objects. For real-world scenes, we use \textbf{COCO-one} and \textbf{COCO-two}
from WhatsUp~\cite{kamath2023s}, which are constructed using images from
MS COCO~\cite{lin2014microsoft}.
COCO-one evaluates the position of a single entity relative to the image,
while COCO-two evaluates the spatial relation between two entities.
Similarly, \textbf{VG-one} and \textbf{VG-two} are constructed from
Visual Genome~\cite{krishna2017visual} and evaluate single-entity and
two-entity spatial reasoning, respectively.
Finally, we use \textbf{CV-Bench-R}, the relation subset of
CV-Bench~\cite{tong2024cambrian}, to evaluate pairwise spatial reasoning
in diverse real-world scenes.

\subsection{Experimental Results}
\label{sec:experimental_results}

As shown in Table~\ref{tab:main_results}, \textsc{INTCORT} achieves the best or tied-best performance on all benchmarks with Qwen-VL-Chat, Qwen2-VL-7B, Qwen2.5-VL-7B, Qwen3-VL-8B, and Molmo-7B. Compared with the five above original model and AdaptVis (the best among all compared methods), \textsc{INTCORT} improves accuracy by $10.01\%$ and $6.47\%$, respectively, averaged across the five models and all seven benchmarks. These results demonstrate that \textsc{INTCORT} achieves significant improvements
in spatial reasoning across diverse vision-language models and benchmarks. When LLaVA-NeXT-7B is equipped with \textsc{INTCORT}, it outperforms AdaptVis on most benchmarks and achieves an average accuracy improvement of $10.52\%$ over the original model across all benchmarks. A possible reason why \textsc{INTCORT} slightly underperforms AdaptVis on a few benchmarks is that it relies on the model's inherent spatial reasoning ability and exploits it through transformed views. Since LLaVA-NeXT encodes relatively weak spatial evidence~\cite{li2026spatialladder}, \textsc{INTCORT} may be less effective, whereas AdaptVis directly adjusts the attention distribution and is less dependent on the model's original spatial reasoning capability. Moreover, as shown in Table~\ref{tab:nonspatial_results}, applying
\textsc{INTCORT} to Qwen2-VL-7B preserves its accuracy on the non-spatial
POPE \cite{li2023evaluating}and MME\cite{fu2026mme} benchmarks, indicating that our method does not compromise the
model's original capabilities on non-spatial tasks.

\subsection{Complementarity Analysis}
\label{sec:complementarity}

% To further demonstrate the benefits of \textsc{INTCORT}, we investigate its
% complementarity with AdaptVis. The two methods are theoretically compatible,
% as \textsc{INTCORT} enhances spatial reasoning through input transformations
% without modifying the VLM's internal mechanisms, while AdaptVis improves
% spatial reasoning by adjusting the model's internal attention distribution. The results presented in Table~\ref{tab:complementarity}. Specifically, among all benchmarks, \textsc{INTCORT} combined with AdaptVis achieves the best results, yielding average improvements of 11.47\%, 24.8\%, and 38.0\% over AdaptVis alone, \textsc{INTCORT} alone, and the original model baseline, respectively. These experimental results validate that our method exhibits strong complementarity with AdaptVis.

To further demonstrate the benefits of \textsc{INTCORT}, we investigate its
complementarity with AdaptVis on LLaVA-NeXT-7B across seven spatial reasoning
benchmarks. The two methods are theoretically compatible,
as \textsc{INTCORT} enhances spatial reasoning through input transformations
without modifying the VLM's internal mechanisms, while AdaptVis improves
spatial reasoning by adjusting the model's internal attention distribution.
The results are presented in Table~\ref{tab:complementarity}. Specifically,
\textsc{INTCORT} combined with AdaptVis achieves the best results, yielding
average improvements of 11.47\%, 24.8\%, and 38.0\% over AdaptVis alone,
\textsc{INTCORT} alone and the original baseline, respectively. These results validate the strong complementarity between our method and AdaptVis.

% \begin{table}[t]
% \centering
% \scriptsize
% \renewcommand{\arraystretch}{0.70}
% \setlength{\aboverulesep}{0pt}
% \setlength{\belowrulesep}{0pt}

% \begin{tabular*}{0.90\columnwidth}{@{\extracolsep{\fill}}lcc@{}}
% \toprule
% Benchmark & Qwen2-VL 7B & +INTCORT \\
% \midrule
% POPE-P & 87.77 & 87.77 \\
% POPE-R  & 88.93 & 88.93 \\
% MME          & 88.08 & \textbf{88.12} \\
% \bottomrule
% \end{tabular*}

% \caption{Accuracy comparison on non-spatial benchmarks.All values are accuracies (\%).}
% \label{tab:nonspatial_results}
% \end{table}

\begin{table}[t]
\centering

{\fontsize{8}{9.2}\selectfont
\renewcommand{\arraystretch}{1.05}
\setlength{\tabcolsep}{7pt}
\setlength{\aboverulesep}{0.7pt}
\setlength{\belowrulesep}{0.7pt}

\begin{tabular}{@{}lcc@{}}
\toprule
Benchmark & Qwen2-VL 7B & +INTCORT \\
\midrule
POPE-P & 87.77 & 87.77 \\
POPE-R & 88.93 & 88.93 \\
MME & 88.08 & \textbf{88.12} \\
\bottomrule
\end{tabular}
}

\caption{Accuracy comparison on non-spatial benchmarks. All values are accuracies (\%).}
\label{tab:nonspatial_results}
\end{table}

\subsection{Ablation}
\label{sec:ablation}
We conduct ablation studies on Qwen2-VL-7B using Controlled-A and VG-one. Specifically, we investigate three variants: replacing the original Qwen2.5-1.5B-Instruct spatial query classifier with Qwen2-0.5B-Instruct, and substituting min confidence aggregation with mean or max aggregation. The results in Table~\ref{tab:ablation} show that all three modifications lead to performance drops, demonstrating the effectiveness of the \textsc{INTCORT} design.

\section{Conclusion}
\label{sec:conc}
This paper proposes \textsc{INTCORT}, a training-free framework for enhancing
the spatial reasoning capability of VLMs. We show that incorrect spatial
predictions from direct inference do not necessarily indicate the absence of
spatial reasoning ability, as alternative inference views generated through
input transformations can recover these failures. By aggregating multi-view
predictions with relation-token confidence, \textsc{INTCORT} improves spatial
reasoning without modifying the VLM's internal mechanisms. Extensive
experiments on six VLMs across seven spatial reasoning benchmarks demonstrate
that \textsc{INTCORT} consistently improves performance over corresponding
base models and existing training-free methods.

\vfill\pagebreak

\bibliographystyle{IEEEbib}
\small
\bibliography{strings,refs}

\end{document}